\documentclass[11pt,a4paper]{article}
\usepackage[hyperref]{naaclhlt2019}
\usepackage{times}
\usepackage{latexsym}
\usepackage{amsmath}
\usepackage{amssymb}
\usepackage{todonotes}
\usepackage{booktabs}

\usepackage{url}

\aclfinalcopy 

\title{Context-Aware Learning for Neural Machine Translation}

\author{S\'ebastien Jean \\
  New York University \\
  {\tt sj2233@nyu.edu} \\\And
  Kyunghyun Cho \\
  New York University \\
  CIFAR Azrieli Global Scholar \\
  {\tt kyunghyun.cho@nyu.edu} \\}

\date{}

\begin{document}
\maketitle
\begin{abstract}

Interest in larger-context neural machine translation, including document-level and multi-modal translation, has been growing. Multiple works have proposed new network architectures or evaluation schemes, but potentially helpful context is still sometimes ignored by larger-context translation models. In this paper, we propose a novel learning algorithm that explicitly encourages a neural translation model to take into account additional context using a multilevel pair-wise ranking loss. We evaluate the proposed learning algorithm with a transformer-based larger-context translation system on document-level translation. By comparing performance using actual and random contexts, we show that a model trained with the proposed algorithm is more sensitive to the additional context.


\end{abstract}

\section{Introduction}

Despite its rapid adoption by academia and industry and its recent success~\citep[see, e.g.,][]{hassan2018achieving}, neural machine translation has been found largely incapable of exploiting additional context other than the current source sentence. This incapability stems from the fact that larger-context machine translation systems tend to ignore additional context, such as previous sentences and associated images. Much of recent efforts have gone into building a novel network architecture that can better exploit additional context however without much success~\citep{D18-1329,gronroos2018memad,laubli2018has}.

In this paper, we approach the problem of larger-context neural machine translation from the perspective of ``learning'' instead. We propose to explicitly encourage the model to exploit additional context by assigning a higher log-probability to a translation paired with a correct context than to that paired with an incorrect one. We design this regularization term to be applied at token, sentence and batch levels to cope with the fact that the benefit from additional context may differ from one level to another. 

Our experiments on document-level translation using a modified transformer~\citep{voita2018context} reveal that the model trained using the proposed learning algorithm is indeed sensitive to the context, contrarily to some previous works~\cite{D18-1329}. We also see a small improvement in terms of overall quality (measured in BLEU). These two observations together suggest that the proposed approach is a promising direction toward building an effective larger-context neural translation model.

\section{Background: Larger-Context Neural Machine Translation}

A larger-context neural machine translation system extends upon the conventional neural machine translation system by incorporating the context $C$, beyond a source sentence $X$, when translating into a sentence $Y$ in the target language. In the case of multimodal machine translation, this additional context is an image which the source sentence $X$ describes. In the case of document-level machine translation, the additional context $C$ may include other sentences in a document in which the source sentence $X$ appears. Such a larger-context neural machine translation system consists of an encoder $f^C$ that encodes the additional context $C$ into a set of vector representations that are combined with those extracted from the source sentence $X$ by the original encoder $f^X$. These vectors are then used by the decoder $g$ to compute the conditional distribution over the target sequences $Y$ in the autoregressive paradigm, i.e.,
\begin{align*}
    p_\theta(y_t | y_{<t}, X, C) = g(y_{<t}, f^X(X), f^C(C)),
\end{align*}
where $\theta$ is a collection of all the parameters in the neural translation model. $f^X$ and $g$ are often implemented as neural networks, such as recurrent networks with attention~\citep{bahdanau2014neural}, convolutional networks~\citep{gehring2017convolutional} and self-attention~\citep{vaswani2017attention}. 

Training is often done by maximizing the log-likelihood given a set of training triplets $\mathcal{D}^{(tr)}=\mathcal{X}^{(tr)}*\mathcal{Y}^{(tr)}*\mathcal{C}^{(tr)}=\left\{ (X_1, Y_1, C_1), \ldots, (X_N, Y_N, C_N)\right\}$. The log-likelihood is defined as
\begin{align}
\label{eq:log-likelihood}
    \mathcal{L}(\theta; \mathcal{D})
    =
    \frac{1}{N}\sum_{n=1}^N
    \sum_{t=1}^{T_n}
    \log p(y_t^n|y_{<t}^n, X_N, C_N).
\end{align}
Once training is done, it is a standard practice to use beam search to find a translation that approximately maximizes
\begin{align*}
    \sum_{t=1}^T \log p(y_t|y_{<t}, X, C).
\end{align*}

\section{Existing approaches to \\~~~~~~~~~~~~~larger-context neural translation}

Existing approaches to larger-context neural machine translation have mostly focused on either modifying the input or the network architecture. \citet{tiedemann2017neural} concatenate the previous source sentence to the current source sentence, which was followed by \citet{bawden2017evaluating} who also concatenate the previous target sentence. \citet{gronroos2018memad} explore various concatenation strategies when the additional context is an image. Other groups have proposed various modifications to the existing neural translation systems~\citep{jean2017does,wang2017exploiting,voita2018context,zhang2018improving,miculicich2018document,maruf2017document,tu2018learning} in the case of document-level translation, while using usual maximum likelihood learning. \citet{zheng2018learning} on the other hand introduces a discriminator that forces the network to improve signal-to-noise ratio in the additional context. In parallel, there have been many proposals on novel network architectures for multi-modal translation~\citep{calixto2017doubly,caglayan2017lium,ma2017osu,libovicky2017attention}. In personalized translation, \citet{michel18acl} bias the output distribution according to the context. All these previous efforts are clearly distinguished from our work in that our approach focuses entirely on a learning algorithm and is agnostic to the underlying network architecture.

\section{Learning to use the context}

In this paper, we focus on ``learning'' rather than a network architecture. Instead of coming up with a new architecture that facilitates larger-context translation, our goal is to come up with a learning algorithm that can be used with any underlying larger-context neural machine translation system.

\subsection{Neutral, useful and harmful context}

To do so, we first notice that by the law of total probability,
\begin{align}
    p_\theta(y_t|y_{<t}, X)
    =& \sum_C p_{\theta} (y_t|y_{<t}, X, C) p(C|X) 
    \nonumber
    \\
\label{eq:independence}
    =& \mathbb{E}_{C\sim C|X} \left[ p_{\theta} (y_t|y_{<t}, X, C) \right]
\end{align}
As such, over the entire distribution of contexts $C$ given a source $X$, the additional context is overall ``neutral''.

When the context $C$ is used, there are two cases. First, the context may be ``useful''. In this case, the model can assign a better probability to a correct target token $y^*_t$ when the context was provided than when it was not:
    \mbox{$p_{\theta}(y^*_t|y_{<t}, X,C) >
    p_{\theta}(y^*_t|y_{<t}, X)$}.
On the other hand, the additional context can certainly be used harmfully:
    \mbox{$p_{\theta}(y^*_t|y_{<t}, X,C) < 
    p_{\theta}(y^*_t|y_{<t}, X)$}.

Although these ``neutral'', ``useful'' and ``harmful'' behaviours are defined at the {\it token level}, we can easily extend them to various levels by defining the following score functions:
\begin{align*}
    &\text{(token) }& s^{\text{tok}}(y_t|\cdot) = \log p_{\theta}(y^*_t|\cdot), \\
    &\text{(sent.) }& s^{\text{sent}}(Y|\cdot) = \sum_{t=1}^T \log p_{\theta}(y^*_t|y_{<t}, \cdot), \\
    &\text{(data) }& s^{\text{data}}(\mathcal{Y}|\cdot) = \sum_{Y \in \mathcal{Y}} s^{\text{sent}}(Y|\cdot).
\end{align*}

\subsection{Context regularization}

With these scores defined at three different levels, we propose to regularize learning to encourage a neural translation system to prefer using the context in a useful way. Our regularization term works at all three levels--tokens, sentences and the entire data-- and is based on a margin ranking loss~\citep{collobert2011natural}:
\begin{align}
\label{eq:context-reg}
    &\mathcal{R}(\theta; \mathcal{D})
    = 
    \\
    \nonumber
    &
    \alpha_d
    \left[
     \left(\sum_{n=1}^N T_n\right)\delta_d - s^{\text{data}}(\mathcal{Y}|\mathcal{X},\mathcal{C}) + s^{\text{data}}(\mathcal{Y}|\mathcal{X})\right]_+
    \\
    \nonumber
    &
    +
    \alpha_s \sum_{n=1}^N
    \left[T_n \delta_s - s^{\text{sent}}(Y_n|X_n, C_n)
    \right.
    \\
    \nonumber
    & \left. \qquad\qquad\qquad\qquad
    + s^{\text{sent}}(Y_n|X_n) \right]_+
    \\
    \nonumber
    &
    +
    \alpha_\tau \sum_{n=1}^N
    \sum_{t=1}^{T_n}
    \left[\delta_\tau - s^{\text{tok}}(y^n_t|y^n_{<t}, X_n, C_n)
    \right.
    \\
    \nonumber
    & \left. \qquad\qquad\qquad\qquad
    + 
    s^{\text{tok}}(y^n_t|y^n_{<t}, X_n)
    \right]_+,
\end{align}
where $\alpha_d$, $\alpha_s$ and $\alpha_\tau$ are the regularization strengths at the data-, sentence- and token-level. $\delta_d$, $\delta_s$ and $\delta_\tau$ are corresponding margin values. 

The proposed regularization term explicitly encourages the usefulness of the additional context at all the levels. We use the margin ranking loss to only lightly bias the model to use the context in a useful way but not necessarily force it to fully rely on the context, as it is expected that most of the necessary information is already contained in the source $X$ and that the additional context $C$ only provides a little complementary information. 

\subsection{Estimating context-less scores}

It is not trivial to compute the score when the context was missing based on Eq.~\eqref{eq:independence}, as it requires (1) the access to $p(C|X)$ and (2) the intractable marginalization over all possible $C$. In this paper, we explore the simplest strategy of approximating $p(C|X)$ with the data distribution of sentences $p_{\text{data}}(C)$.


We assume that the context $C$ is independently distributed from the source $X$, i.e., $p(C|X)=p(C)$ and that the context $C$ follows the data distribution. This allows us to approximate the expectation by uniformly selecting $M$ training contexts at random:
\begin{align*}
    s(\cdot|\cdot) = \log p(\cdot|\cdot) \approx
    \log \frac{1}{M} \sum_{m=1}^M p(\cdot|\cdot, C_m),
\end{align*}
where $C^m$ is the $m$-th sample. 

A better estimation of $p(C|X)$ is certainly possible. One such approach would be to use a larger-context recurrent language model by \citet{wang2015larger}. Another possible approach is to use an off-the-shelf retrieval engine to build a non-parametric sampler. We leave the investigation of these alternatives to the future.




\subsection{An intrinsic evaluation metric}

The conditions for ``neutral'', ``useful'' and ``harmful'' context also serve as bases on which we can build an intrinsic evaluation metric of a larger-context neural machine translation system. We propose this metric by observing that,
for a well-trained larger-context translation system,

\begin{align*}
    \Delta^{\mathcal{D}}(\theta) = s(\mathcal{Y}|\mathcal{X},\mathcal{C}; \theta) - s(\mathcal{Y}|\mathcal{X}; \theta) > 0,
\end{align*}
while it would be 0 for a larger-context model that completely ignores the additional context. We compute this metric over the validation set using the sample-based approximation scheme from above. Alternatively, we may compute the difference in BLEU ($\Delta_{BLEU}^{\mathcal{D}}(\theta)$) over the validation or test data. These metrics are complementary to others that evaluate specific discourse phenomena on specially designed test sets~\cite{bawden2017evaluating}.



\begin{table*}[t]
\begin{center}
\begin{tabular}{ccc||ccc}
& & & \multicolumn{2}{c}{BLEU} \\
& Context & Context-Aware Reg. & Normal & Context-Marginalized & $\Delta_{BLEU}^{\mathcal{D}^{test}}(\theta)$\\
\toprule
(a) & $\circ$  & $\circ$ & 29.16 (29.62) & - & - \\
(b) & $\circ^\dagger$ &$\circ$ & 29.23 (29.65) & 29.23 (29.65) & 0\\
\midrule
(c) & $\bullet$ &$\circ$ & 29.34 (29.63) & 28.94 (29.23) & 0.40\\
\midrule
(d) & $\bullet$ & $\bullet$ & 29.91 (30.13) & 26.17 (25.82) & 3.74 \\
\end{tabular}
\end{center}

\vspace{-3mm}
\caption{We report the BLEU scores with the correctly paired context as well as with the incorrectly paired context (context-marginalized).  Context-marginalized BLEU scores are averaged over three randomly selected contexts. BLEU scores on the validation set are presented within parentheses. $\dagger$ Instead of omitting the context, we give a random context to make the number of parameters match with the larger-context model.}
\label{tbl:results}

\vspace{-4mm}
\end{table*}

\section{Experimental Settings}

\paragraph{Data}

We use En$\rightarrow$Ru parallel data from OpenSubtitles2018~\cite{lison2018opensubtitles} and choose the same training data subset of 2M examples as~\cite{voita2018context} did. We build a joint vocabulary of BPE subword tokens between the source and target languages using 32k merge operations~\cite{sennrich2015neural}.

\paragraph{Context-less score estimation}

We simply shuffle the context in each minibatch to create $M=1$ random context per example.



\begin{figure}
\begin{center}
\includegraphics[width=.8\columnwidth]{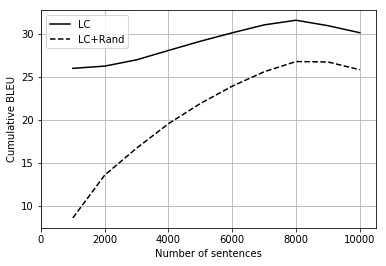}
\end{center}
\vspace{-4mm}
\caption{Cumulative BLEU scores on the validation set sorted by the sentence-level score difference according to the larger-context model.}
\label{fig:cumul}
\vspace{-6mm}
\end{figure}

\paragraph{Models}

We build a larger-context variant of the base transformer~\citep{vaswani2017attention} that takes as input both the current and previous sentences, similarly to that by \citet{voita2018context}. Each of the current and previous sentences is independently encoded by a common 6-layer transformer encoder. The final representation of each token in the current sentence is obtained by attending over the final token representations from the past sentence and combining the outputs from the current and past sentences nonlinearly. We use the same decoder from a standard transformer, and share all the word embedding matrices. See the appendix for the detailed description. 

We use Adam with an initial step size of $10^{-4}$ to train each model. We evaluate the model every half epoch using greedy decoding and halve the learning rate when the BLEU score on the development does not improve for five consecutive evaluations, following \cite{denkowski2017stronger}. Based on the BLEU score on the validation set during the preliminary experiments, we set the coefficients and margins of the proposed regularization term \eqref{eq:context-reg} to  $\alpha_\tau=\alpha_d=1$, $\alpha_s=0$, $\delta_\tau=\delta_s=0$ and $\delta_d=\text{log}(1.1)$. Models are evaluated with a beam size of 5, adjusting scores according to length~\cite{wu2016google}.

\section{Result and Analysis}

In Table~\ref{tbl:results}, we present the translation quality (in BLEU) of the four variants. We make a number of observations. First, the use of previous sentence (c) does not improve over the baseline (a--b) when the larger-context model was trained only to maximize the log-likelihood \eqref{eq:log-likelihood}. We furthermore see that the translation quality of the larger-context model only marginally degrades even when the incorrectly paired previous sentence was given instead ($\Delta_{BLEU}^{\mathcal{D}^{test}}(\theta) = 0.40$), implying that this model largely ignores the previous sentence.

Second, we observe that the larger-context model improves upon the baselines, trained either without any additional context (a) or with purely random context (b), when it was trained with the proposed regularization term (d). The evaluation metric $\Delta_{BLEU}^{\mathcal{D}^{test}}(\theta)$ is also significantly larger than $0$, suggesting the effectiveness of the proposed regularization term in encouraging the model to focus on the additional context.

In Fig.~\ref{fig:cumul}, we contrast the translation qualities (measured in BLEU) between having the correctly paired (LC) and incorrectly paired (LC+Rand) previous sentences. The sentences in the validation set were sorted according to the difference $s^{\text{sent}}(Y|X,C) - s^{\text{sent}}(Y|X)$, and we report the cumulative BLEU scores. The gap is large for those sentences that were deemed by the larger-context model to benefit from the additional context. This match between the score difference (which uses the reference translation) and the actual translation quality further confirms the validity of the proposed approach.





\section{Conclusion}

We proposed a novel regularization term for encouraging a larger-context machine translation model to focus more on the additional context using a multi-level pair-wise ranking loss. The proposed learning approach is generally applicable to any network architecture. Our empirical evaluation demonstrates that a larger-context translation model trained by the proposed approach indeed becomes more sensitive to the additional context and outperforms a context-less baseline. We believe this work is an encouraging first step toward developing a better context-aware learning algorithm for larger-context machine translation. We identify three future directions; (1) a better context distribution $p(C|X)$, (2) efficient evaluation of the context-less scores, and (3) evaluation using other tasks, such as multi-modal translation.







\section*{Acknowledgments}

SJ thanks NSERC. KC thanks support by AdeptMind, eBay, TenCent, NVIDIA and CIFAR. This work was partly supported by Samsung Advanced Institute of Technology (Next Generation Deep Learning: from pattern recognition to AI) and Samsung Electronics (Improving Deep Learning using Latent Structure).

\bibliography{naaclhlt2019}
\bibliographystyle{acl_natbib}

\clearpage
\appendix

\section{Larger-Context Transformer}

A shared 6-layer transformer encoder is used to independently encode an additional context $C$ and a source sentence $X$.

\begin{align*}
  c &= \text{TransformerEnc}_6(C) \\
  x &= \text{TransformerEnc}_6(X)
\end{align*}

Using $x$ as queries ($q$), a multi-head attention mechanism attends to $c$ as key-values ($k,v$).
The input and output are merged through a gate.\footnote{Current gate values are unbounded, but it may be preferable to apply a sigmoid function to restrict the range between 0 and 1.} The final source representation is obtained through a feed-forward module (FF) used in typical transformer layers.

\begin{align*}
  \hat{x}_c &= \text{Attn}(q=x; k,v=c) \\
  g &= \text{Linear}([x;\hat{x}_c]) \\
  x_c &= \text{FF}(g\cdot \text{Dropout}(\hat{x}_c) + (1-g)\cdot x)
\end{align*}

We use a standard 6-layer transformer decoder which attends to $x_c$.

\begin{equation*}
      p_\theta(y_t | y_{<t}, X, C) = \text{TransformerDec}_6(y_{<t}, x_c),
\end{equation*}

\end{document}